# On Statistical Learning of Branch and Bound for Vehicle Routing Optimization

Andrew Naguib[a,*], Waleed A. Yousef[a], Issa Traoré[a], Mohammad Mamun[b]

[a]*Department of Electrical and Computer Engineering, University of Victoria, 3800 Finnerty Road, Victoria, V8P 5C2, BC, Canada*
[b]*National Research Council, Fredericton, NB, Canada*

**Abstract**

Recently, machine learning of the branch and bound algorithm has shown promise in approximating competent solutions to NP-hard problems. In this paper, we utilize and comprehensively compare the outcomes of three neural networks—graph convolutional neural network, GraphSAGE, and graph attention network—to solve the capacitated vehicle routing problem (CVRP). We train these neural networks to emulate the decision-making process of the computationally expensive Strong Branching strategy. The neural networks are trained on six instances with distinct topologies from the CVRPLIB and evaluated on eight additional instances. Moreover, we reduced the minimum number of vehicles required to solve a CVRP instance to a bin-packing problem, which was addressed in a similar manner. Through rigorous experimentation, we found that this approach can match or improve upon the performance of the branch and bound algorithm with the Strong Branching strategy while requiring significantly less computational time, even in comparison with the heuristic strategy referred to as the Reliable Pseudo-cost. The source code corresponding to our research findings and methodology is accessible and available for reference at the following web address: https://isotlaboratory.github.io/ml4vrp.

*Keywords:* branch and bound, combinatorial optimization, machine learning, transportation, logistics

## 1. Introduction

The capacitated vehicle routing problem (CVRP) (Dantzig and Ramser 1959) is a well-studied combinatorial optimization challenge with applications across diverse domains. It centers on the strategic allocation of a fleet of vehicles, all stationed at a common depot, to a set of customers with distinct demands. The primary objective is to facilitate a configuration of vehicle routes that minimizes the overall operational cost. When the vehicle fleet consists of a single unit, the CVRP reduces to the traveling salesperson problem (TSP).

Our study is focused on approximating solutions to the CVRP using graph neural networks (we shall expound on the details later in this section). Nevertheless, we would like to emphasize on the interrelation between the CVRP and the bin packing problem (BPP), a classic combinatorial problem in its own right. Remarkably, in situations where the travel expenses between customers are negligible (i.e., zero), the CVRP and BPP become equivalent. Furthermore, an intriguing parallel can be drawn between computing the minimal vehicle count required to attain feasible solutions within a CVRP instance and solving a bin packing problem–the primary reason for its integration into our work, which will obviate the need to "guess" the number or using a lower bound which is not necessarily feasible.

The CVRP and BPP are NP-hard problems (Lenstra and Kan 1981; Martello and Toth 1990). Exact solutions and approximate ones (Arora and Barak 2009; Papadimitriou and Yannakakis 1988) cannot be computed *efficiently* unless the P = NP conjecture holds true. Heuristics with bounded-error ratios (Uchoa et al. 2017) are alternatively often used. Instead of using the general-purpose strategies, which are subject to performance limitations (Ho and Pepyne 2001), we spur the specialization of deep neural networks (especially the geometric ones) to specifically solve the CVRP and BPP rather than training an omniscient one. Statistical models possesses the advantage of having a constant inference complexity. Moreover, the training time required for solving the CVRP or BPP is, as we shall demonstrate, negligible compared to the time needed to solve either deterministically.

We substitute the branching strategy of the branch and bound algorithm (B&B) (Land and Doig 1960) with one out of three graph neural networks. Then, we execute the solver and compare the duality gap between the native and trained branching strategies. To strengthen our analysis, we investigated the trained neural networks in a deterministically measurable environment, the mathematical solver SCIP (Bestuzheva et al. 2021).

The neural networks are trained on annotated samples of the decisions made by the strong branching strategy (SB) and compared against the reliable pseudo-cost strategy (RPC) (Applegate et al. 1995; Achterberg, Koch, and Martin 2005). The SB strategy (although computationally expensive) is guaranteed to obtain the optimum solution in the smallest B&B tree size. Hence, its decisions are efficient for training. Conversely,

*Corresponding author
*Email addresses:* ndrwnaguib@gmail.com (Andrew Naguib), wyousef@uvic.ca (Waleed A. Yousef), itraore@ece.uvic.ca (Issa Traoré), Mohammad.Mamun@nrc-cnrc.gc.ca (Mohammad Mamun)

the RPC branching strategy finds a suboptimal but high-quality solution in considerably less time, making it an ideal baseline for assessing temporal efficiency. We describe the intricacies of B&B algorithm as well as the SB and RPC strategies in Sections 2.3 and 2.4.

The combinatorial nature of the CVRP or BPP allows for numerous mathematical formulations. In this context, we adopt an integer programming perspective, which raises three fundamental questions:

(A) How can graph neural networks be trained to approximate solutions for CVRP and BPP using an integer programming framework?

(B) What trade-offs exist among solution quality, runtime, and training sample complexity?

(C) How can we evaluate the generalization error of these approaches across various graph topologies that may differ from those encountered during training?

In response to question (A), we have devised two graph neural networks employing the architectures of the graph attention network (GAT) (Veličković et al. 2017) and the GraphSAGE model (Hamilton, Ying, and Leskovec 2017), while also assessing the graph convolutional neural network model (GCNN) proposed by (Gasse et al. 2019). Each of the three networks was trained on decision samples derived from the SB strategy while solving instances of the CVRP from the CVRPLIB (Lima et al. 2014) and the BPP from the Operation Research Library (Beasley 2004). We generated the decision samples by developing a fork of the École framework (Prouvost et al. 2020). Internally, École hooks events' listeners on the solving process taking place inside of SCIP, the underlying mathematical solver.

As for question (B), we evaluated the efficacy of the classifiers established in (A) compared to the SB and RPC branching strategies. By conducting eight hundred experiments, we found that the classifiers consistently produced solutions that were either equivalent to, frequently superior to, or within a small fraction of the precision of the solutions derived from the SB or RPC, and in 2x-8x less time.

The challenge in (C) relates to delineating the concept of generalization error in our setup, which is distinct from the traditional approach of assessing the error using a sequestered test set. The accurate prediction of the branching variables in a test set by the classifiers does not unambiguously imply its ability to make valid decisions in instances of the CVRP with complex topologies. Thus, the question at hand is more challenging: How to measure the complexity of an integer program? Several factors require consideration, such as the instance topology, nature of the constraints, and number of binary variables. These factors also influence the estimated tree-size (TSE) of the B&B algorithm (Hendel et al. 2022; Özaltın, Hunsaker, and Schaefer 2011; Cornuéjols, Karamanov, and Li 2006), which we report in alongside with the graph edit distance (GED) (Abu-Aisheh et al. 2015) for each training and evaluation instance pair. We aim to establish a credible relation between the quality of solutions and complexity of the problem.

The rest of the manuscript is organized into different sections, beginning with Section 2, which provides an overview of the pertinent concepts employed in our research, followed by a literature review on the CVRP in Section 3. We elaborate on our approach in Section 4. We describe the experimental design in Section 5 and present the study's results in Section 6. Finally, in Section 7, we discuss these results, the advantages and limitations of our method, and suggest potential avenues for future research.

## 2. Background

### *2.1. Integer Program*

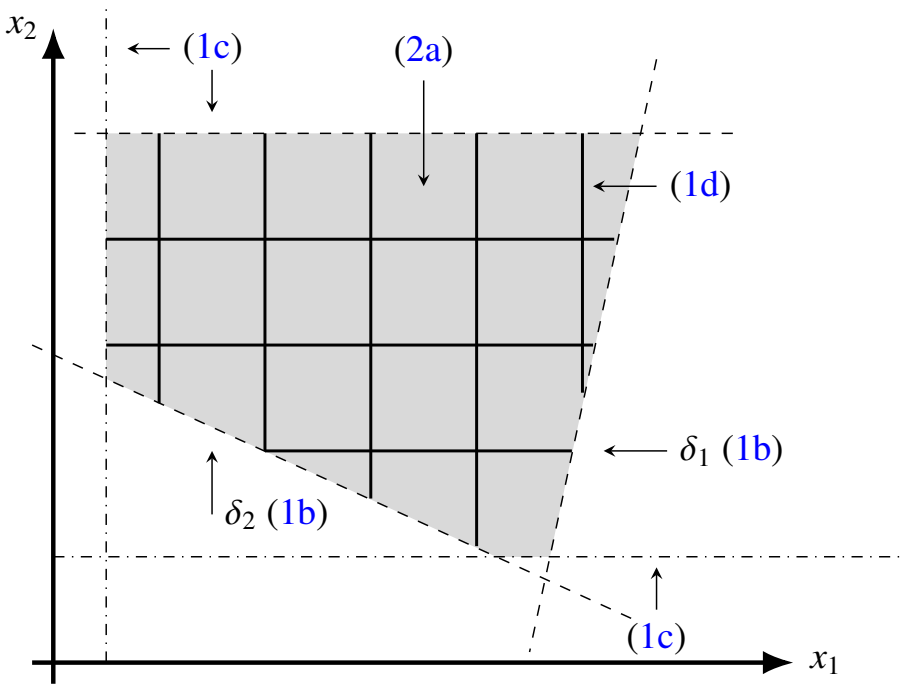

Figure 1: An IP example consisting of two variables and three constraints.

Let $x, c, l, u \in \mathbb{R}^n$, $A \in \mathbb{R}^{m \times n}$, $b \in \mathbb{R}^m$, $I \subseteq \{1, 2, \cdots, n\}$, and $l, u \in \mathbb{R}^n$. An integer program (IP) can be defined as follows:

$$\underset{x}{\text{minimize}} \quad c^T x \tag{1a}$$
$$\text{subject to} \quad Ax \leq b, \tag{1b}$$
$$l \leq x \leq u, \tag{1c}$$
$$x_i \in \mathbb{Z} \quad i \in I. \tag{1d}$$

A *feasible* assignment $x_i = z_i$ must satisfy the above constraints (1b)-(1d), and an *optimal* one $z^\star$, called the *minimizer*, is both feasible and minimizes the objective (1a). For a shorthand notation, let $\delta_i$ refer to $A_i x_i \leq b_i$ in (1b).

The *linear programming relaxation* (LP-relaxation) for the IP defined above would be:

$$\underset{x}{\text{minimize}} \quad c^T x \tag{2a}$$
$$\text{subject to} \quad Ax \leq b, \tag{2b}$$
$$l \leq x \leq u. \tag{2c}$$

That is, the integrality constraints are *relaxed*, which also makes the problem convex. The optimal solution to (2a) is also optimal to the original problem (1a) if and only if it satisfies (1d).

Figure 1 is an example of an integer program consisting of two variables, $\{x_1, x_2\}$, and two constraints, $\{\delta_1, \delta_2\}$. The gray shading represents the relaxed problem, (2a), where the solid grid lines are the integrality constraints, (1d), and the dashed lines, (1c), are the box constraints.

### 2.2. Primal-Dual Method

The quality of each feasible solution found to (1a) is assessed using the primal-dual method. The *primal (upper) bounds* $p^\star$ are provided by feasible solutions, and the *dual (lower) bounds* $d^\star$ by *relaxation* or duality. The Lagrangian dual problem (S. P. Boyd 2004 - 2004) to (2a) is:

$$\underset{\lambda}{\text{maximize}} \quad -b^T\lambda \tag{3a}$$
$$\text{subject to} \quad A^T\lambda + c \geq 0, \tag{3b}$$

where $\lambda$ is the Lagrange multiplier or the dual variable for the inequality constraints (2b). The dual problem acts as a certificate on the limit of the performance, i.e., the upper bound that declares optimality of $p^\star$. The duality *gap* $f$ is given by $p^\star - d^\star$. A gap $f > 0$ or $f = 0$ indicates a *weak-* or *strong-duality*, respectively. We define the relative gap to be:

$$\Delta f = \frac{|p^\star - d^\star|}{min\{|p^\star|, |d^\star|\}}.$$

In sequel, we shall drop the term "relative" and refer to $\Delta f$ as gap.

### 2.3. Branch and Bound

The two building blocks in the algorithm are branching and bounding. In the branching step, the problem is divided into several smaller and less constrained ones and the bounding step selects which subproblems to solve next.

Formally, for a set of variables, $\{x_1, x_2, \cdots, x_k\}$, the algorithm initializes the list of branching candidates $S = \{x_1\}$ (i.e., the first node), the optimal assignments $z_i^\star = \phi$, and the optimal gap $f^\star = -\infty$ (or any other heuristically known and feasible lower bound).

Then, (1) an LP-relaxation, $\text{LP}_i$ (2a), is solved for the candidates in $S$ *(bounding)*, after which they are removed. (2) If the $\text{LP}_i$ yields an infeasible solution $z_i \notin [l_i, u_i]$ and $S = \{\}$ (i.e., no more branching candidates), the algorithm halts. Otherwise, (3) another feasibility check is made to the *original* integer program (1a)-(1d), if $z_i \in \mathbb{Z}$ (1d), then $z_i^\star = z_i$ and $f^\star = f_{\text{LP}_i}$, which would then be an optimal solution. If $z_i \notin \mathbb{Z}$, (4) then sub linear programs $\text{LP}_{i0}, \cdots, \text{LP}_{in}$ are constructed from the current $\text{LP}_i$ *(branching)* and whose union of the feasible-solution space does not contain $z_i$.

There are several strategies to drive the behavior of the branching and bounding steps (Morrison et al. 2016; Wosley 2020). A lower bound has not yet been proven for the algorithm and still an open question (Lipton and Regan 2012). The time required to solve a problem increases *exponentially* with the number of variables.

### 2.4. Strong Branching (SB) and Reliable Pseudo-Cost (RPC) Branching Strategies

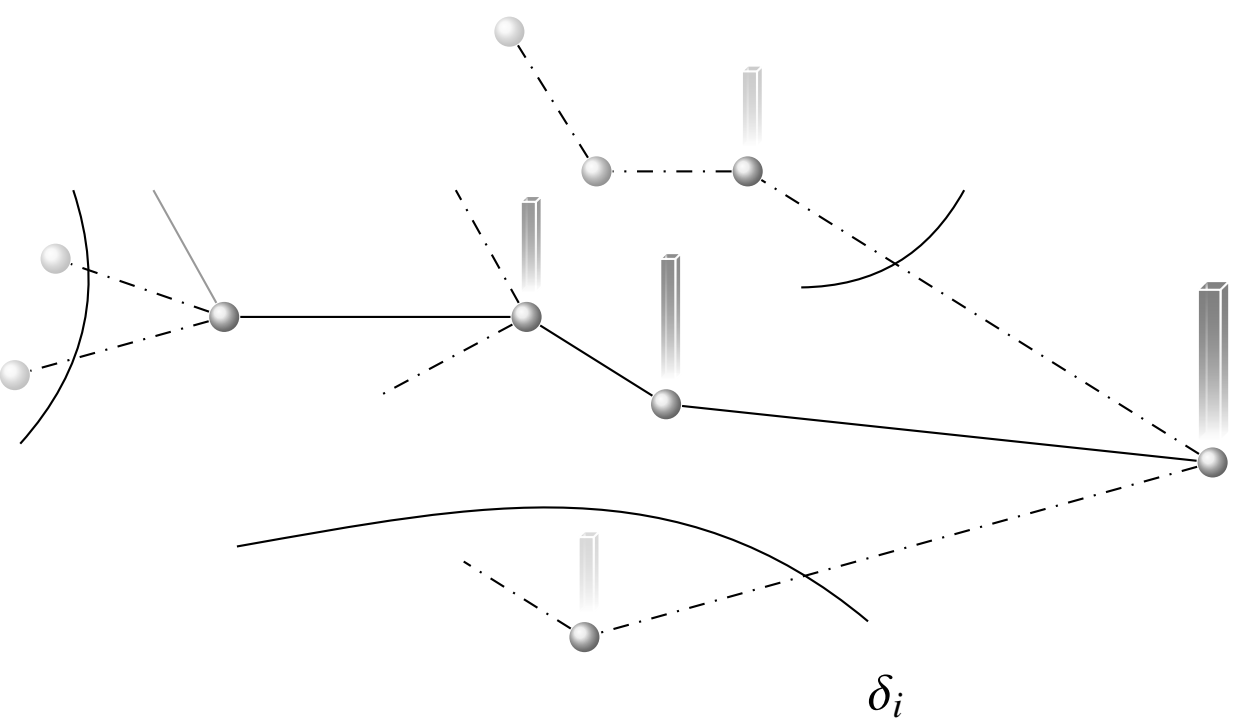

Figure 2: Illustration of the branch and bound process. The *(pruned)* dashed branches are depicted with arcs corresponding to a constraint $\delta_i$. The superimposed bars represent the upper bounds.

The crux of SB is to carefully branch the tree to guarantee the smallest B&B tree size by performing a one-step look-ahead before deciding to branch. The solver starts by choosing a set of integer variables, $S$, that are fractional in the LP-relaxation (when $S$ represents all the integer variables, the strategy is called the Full Strong Branching). Then, temporal lower and upper bounds are calculated for the selected variable and upon which the subsequent paths are decided. In Figure 2, you can see a visual representation of the process. Each spherical node represents a variable, and the hovering bar shows its upper bound. The constraints, $\delta_i$, play a role in deforesting the solution tree.

The SB strategy is commonly used in conjunction with other branching strategies to balance the trade-off between the solution quality and computational cost. On the other hand, the RPC branching strategy avoids the situation by assigning an estimated cost to each variable based on the results of the previous subproblems while occasionally using the SB strategy on the *unreliable* pseudo-costs according to a predefined reliability constant.

### 2.5. Capacitated Vehicle Routing Problem (CVRP)

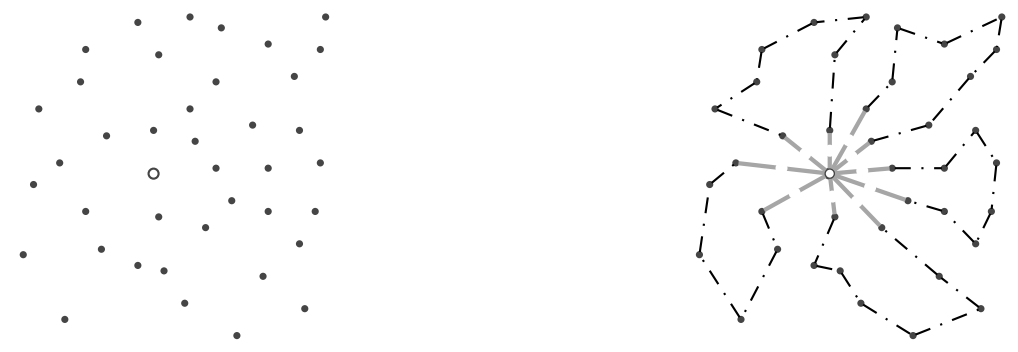

(a) (b)

Figure 3: A CVRP example with $n = 40$ and $k = 5$. Given the dispersed points shown in (a), CVRP asks to find the routes presented in (b).

Let $G(V, \vec{A})$ be a *complete* graph consisting of the vertex set $V = \{0\} \cup N$ where $N = \{1, 2, \cdots, n\}$ and the *directed* arcs set $\vec{A} = \{(i, j) : (i, j) \in V \times V, i \neq j\}$. There is a cost $c_{ij}$ incurred for using an arc $a_{ij} \in \vec{A}$. For simplicity, we will assume that $c_{ij} = c_{ji}$. Each customer $i \in N$ has a nonnegative demand $q_i$.

The *out-degree* of a node is denoted by $\deg^+(i)$ for $i \in V$, that is, the number of arcs $\vec{a}_{ij}$ *leaving* from node $i$ to node $j \in V \backslash \{i\}$. Similarly, the *in-degree* of a node is denoted $\deg^-(v)$. At the depot $\{0\}$, there is a fleet of vehicles of size $k$ with identical capacities $Q$. The goal is to construct a tour $\vec{T} \subseteq \vec{A}$ for each vehicle where

(i) Each customer $i \in N$ is visited exactly once and by a single vehicle or tour $T$;

(ii) The demand served by tour $\vec{T}$ does not exceed the capacity $Q$;

(iii) Each tour $\vec{T}$ start and finish at the depot (to eliminate subtours);

(iv) The constructed tours jointly minimize the total cost and serve the total demand $\sum_{i=1}^{n} q_i$.

The corresponding integer program is:

$$
\begin{aligned}
\underset{x}{\text{minimize}} \quad & \sum_{(i,j)\in\vec{A}} c_{ij}x_{ij} && && \text{(4a)}\\
\text{subject to} \quad & \sum_{j\in\deg^+(i)} x_{ij} = 1 && \forall\, i \in V, && \text{(4b)}\\
& \sum_{i\in\deg^-(i)} x_{ij} = 1 && \forall\, j \in V, && \text{(4c)}\\
& \sum_{j\in\deg^+(0)} x_{0j} = k, && && \text{(4d)}\\
& \sum_{j\in\deg^-(0)} x_{0j} = k, && && \text{(4e)}\\
& u_i - u_j + Qx_{ij} \le Q - q_j && \forall\, (i,j) \in \vec{T}, && \text{(4f)}\\
& q_i \le u_i \le Q && \forall\, i \in N, && \text{(4g)}\\
& x_{ij} \in \{0,1\} && \forall\, (i,j) \in \vec{A}, && \text{(4h)}
\end{aligned}
$$

where $x \in \mathbb{Z}^{|V|\times|V|}$ and $u \in \mathbb{R}^{|\vec{T}|}$ are the decision variables representing the arc set $\vec{A}$ and constructed tour $\vec{T}$, respectively. The equations (4b) and (4c) stipulate constraint (i). The subtour elimination and capacity constraints (SEC), which correspond to constraints (ii) and (iii), are guaranteed by (4f) and (4g). These equations are the Miller-Tucker-Zemlin (MTZ)-formulation of SEC (C. E. Miller, Tucker, and Zemlin 1960). The integrality constraints are enforced by equation (4h).

### 2.6. Bin Packing Problem (BPP)

Formulating an IP for the BPP in terms of the CVRP can be written by removing constraints (4f)-(4g) and through a simple algebraic manipulation to minimize the number of vehicles, referred to as bins in this context, in lieu of (4a). We achieve so by expanding the value $k$ to the *bins* variable $K = \{1, 2, \cdots, k\}$, and restricting that the demands $q_j \forall j \in N$, represented as weighted items, must be distributed across the available bins without exceeding the bin's capacity $Q$.

$$
\begin{aligned}
\underset{k}{\text{minimize}} \quad & \sum_{i}^{k} k_i && && \text{(5a)}\\
\text{subject to} \quad & \sum_{i}^{k} q_j x_{ij} \le Qk_i && \forall\, j \in N, && \text{(5b)}\\
& \sum_{i}^{k} x_{ij} = 1 && \forall\, j \in N, && \text{(5c)}\\
& k_i \in \{0,1\} && \forall\, i \in K, && \text{(5d)}\\
& x_{ij} \in \{0,1\} && \forall\, i \in K, j \in N. && \text{(5e)}
\end{aligned}
$$

The decision variable $x$ represents the allocation of customers' demands or items to vehicles or bins, with each entry $x_{i,j}$ representing the assignment of the $i$-th vehicle or bin to the $j$-th demand or item.

We reformat equation (5b) as $\sum_i^k q_j x_{ij} - Qk_i \le 0$ to match the format required by a SCIP constraint. The solution to this integer program, let it be denoted $k_{\min}$, replaces the value of $k$ in (4f)-(4g), which is frequently set to be the strict lower bound calculated as $\lceil \frac{\sum_i^n q_i}{Q} \rceil$–however, this lower bound is not necessarily feasible. The equation(s) (5b) represents the bin capacity constraint, (5c) ensures that each item is assigned to a single bin, and (5d)-(5e) are the integrality constraints.

## 3. Literature Review

Solving optimization problems by using neural networks emerged from the work of (Hopfield and Tank 1985), who formulated the TSP as an energy minimization problem and showed that the energy function decreases as the algorithm (a neural network) progresses to a local optimum; that said, it was still behind several heuristics. Attempts to utilize the same idea for CVRP were first examined by (Torki, Somhon, and Enkawa 1997; Ghaziri 1996). Although the internals of the two approaches differ, they both employ self-organizing feature maps and show on-par performance with other heuristics such as (Clarke and Wright 1964; Gillett and L. R. Miller 1974). Building a search tree based on the branch and bound algorithm has remained a promising approach for decades and many studies have contributed to guiding heuristics and branching policies.

The effectiveness of using a machine learning model to imitate the decisions of the B&B algorithm by replacing the underlying branching policy was first studied by (Khalil et al. 2016). They trained a surrogate function to rank the decisions collected from the SB branching policy, treating it as a learning-to-rank problem. Their results showed promise for this approach. Later, (Gasse et al. 2019) used deep neural networks to further improve this idea, formulating the B&B as a Markov decision process and evaluating it on four NP-hard problems: set cover, capacitated facility location, combinatorial auction, and maximum independent set. Notably, (Nair et al. 2020) were able to build on this work and develop a batch LP-solver that generates 2.6 times more data samples than its sequential counterpart–the underlying theoretical framework is alternating direction method of multipliers (S. Boyd et al. 2011), which breaks an optimization problem into sum of two or more simpler functions.

Another method was provided by (Zarpellon et al. 2020), who handcrafted input features with an improved signal-to-noise ratio by adding a richer context of the variable selection process (for instance, adding the variable participation history in the search process and previous branches). The authors also contributed two novel deep neural network architectures and empirically showed that the GCNN framework proposed by (Gasse et al. 2019) poorly generalizes to generic MIP problems compared to their model. Since modern solvers are inherently CPU-based, (Gupta et al. 2020) investigated the possibility of using a GNN (which is GPU-intensive) only at the B&B root node, in which the rest of the tree is explored using the simple multi-layer perceptron.

The inception of end-to-end learning for optimization problems occurred just seven years ago when (Vinyals, Fortunato, and Jaitly 2015) introduced the Pointer Network. This network provided suboptimal solutions to three geometric tasks, one of which is the TSP. The authors initiated the approach by solving the problem (up to 20 nodes) using the Held-Karp algorithm (Bellman 1962), which has a time complexity of $O(2^n n^2)$, and generating a labeled training set–solving larger instances was non-viable due to high computational costs. The approach was further extended by (Bello et al. 2016) to adopt a reinforcement learning approach, predicting the distribution of possible route permutations (similar to having a value function) and guiding the agent using the negative tour length as the reward signal.

For CVRP, (Nazari et al. 2018) used the original Pointer Networks without the encoding step, which uses a recurrent neural network, arguing that it adds a layer of complexity (since there is no sequential information in TSP) and that the model generalizes to larger problems without it. Investigations into learning heuristic methods were also conducted by (Kool, van Hoof, and Welling 2018), who proposed an attention-based encoder-decoder architecture trained using REINFORCE with baseline (Sutton and Barto 2018), their architecture expects graph nodes (e.g., customers, cities, etc.) as input and outputs a permutation of those nodes (the optimal tour found). Despite being designed for TSP, one can modify the architecture to work on other problems.

For a comprehensive study of machine learning approaches for combinatorial optimization, we refer the reader to (Bengio, Lodi, and Prouvost 2021). Additionally, (paolo toth paolo and daniele vigo daniele 2002; Martello and Toth 1990) can be regarded as the authoritative sources for CVRP and BPP, respectively.

Our research extends upon the theoretical framework proposed by (Khalil et al. 2016; Gasse et al. 2019; Nair et al. 2020) as a foundation for solving the CVRP. We additionally encompass the estimation of the minimum requisite number of vehicles, a facet achieved by leveraging solutions for the BPP. Two of the three deep graph neural networks utilized in our approach have not been previously employed to learn to branch and/or bound, and all three networks remain novel to solving the CVRP and BPP. Moreover, we scrutinized the generalization error of our approach concerning those two problems. We conducted a rich empirical analysis that included numerous topological structures to confirm similarly learned patterns and refute incongruous ones.

## 4. Method

We model each branching step, as described in Section 2.3, using a bipartite node with two disjoint sets. These sets, denoted as $x = \{x_1, x_2, \cdots, x_n\}$ and $\delta = \{\delta_1, \delta_2, \cdots, \delta_m\}$ (see Figure 4), represent the variables and constraints in (1b), respectively. The sensitivity of a variable to the objective function (either (4a) for the CVRP or (5a) for the BPP) is represented by an edge weight, or coefficient, $a_{ij} \in A$, where $i$ belongs to Set $x$ and $j$ belongs to Set $\delta$. Each element in either set has a feature vector that describes its attributes during the solving process. For example, the feature vector for a variable $i \in x$ might include the objective value, variable type (e.g., binary, integer, or continuous), and lower and upper bounds. Similarly, the feature vector for a constraint $j \in \delta$ may include the tightness, dual problem solution value (3a), scaled age, and bias. In total, there are 19 features collected for each variable and 5 for each constraint.

$$
\begin{array}{lllllll}
\underset{x}{\text{minimize}} & c_1 x_1 & + c_2 x_2 & + \cdots + & c_n x_n & & \\
\text{subject to} & a_{11} x_1 & + a_{12} x_2 & + \cdots + & a_{1m} x_n & \leq & b_1 \\
 & \vdots & & \ddots & \vdots & & \vdots \\
 & a_{m1} x_1 & + a_{m2} x_2 & + \cdots + & a_{mn} x_n & \leq & b_m
\end{array}
$$

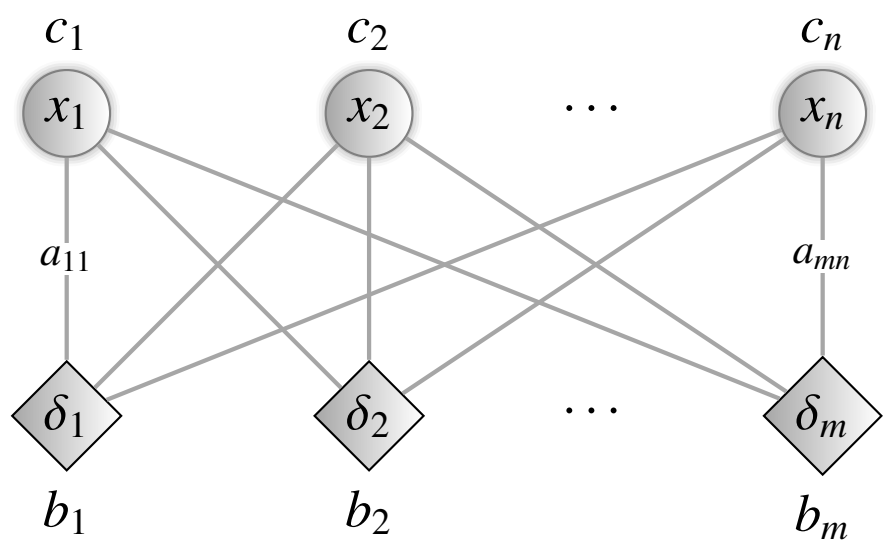

Figure 4: portraying equations (1a)-(1b) as a bipartite graph; the corresponding integer program would then be $\underset{\circ}{\text{minimize}} \sum \circ$; subject to $\diamond$

We describe the three deep graph neural networks—GCNN, GraphSAGE, and GAT—using the message-passing protocol (Hamilton 2020). Although the networks may differ in how we implement the AGGREGATE and UPDATE operations, they operate on the same basic principle. First, we aggregate a *message*, the features of the node's local neighborhood, and update the node's feature vector using some binary operation with the node's current feature vector and the aggregated message being its operands.

### *4.1. Graph Convolutional Neural Network (GCNN)*

In the GCNN designed by (Gasse et al. 2019), the input is represented by a triplet $(\delta, A, x)$. The first layer transforms the elements into latent vectors $H_\delta \in \mathbb{R}^{m\times d}$, $H_A \in \mathbb{R}^{m\times n\times d}$, and $H_x \in \mathbb{R}^{n\times d}$, where $d$ is the embedding dimension.

The transformation module for $\delta$ and $x$, includes a layer normalization (LayerNorm, as proposed by (J. L. Ba, Kiros, and Hinton 2016)), linear transformation, and rectified linear unit (ReLU) as the nonlinearity, which we denote as $\sigma$. For convenience, we construct $f$ to embed the elements $i \in H_x$ and $j \in H_\delta$ as follows:

$$f(y) = \begin{cases} y, & \text{if } l = 0 \\ \sigma\left(W^{(l)}\text{LayerNorm}(y)\right), & \text{otherwise} \end{cases} \tag{6}$$

$$H_x^{(0)} = \{\underbrace{f(x_1)}_{h_{x_1}^{(0)}}, \underbrace{f(x_2)}_{h_{x_2}^{(0)}}, \cdots, \underbrace{f(x_n)}_{h_{xn}^{(0)}}\}, \tag{7}$$

$$H_\delta^{(0)} = \{\underbrace{f(\delta_1)}_{h_{\delta_1}^{(0)}}, \underbrace{f(\delta_2)}_{h_{\delta_2}^{(0)}}, \cdots, \underbrace{f(\delta_m)}_{h_{\delta_m}^{(0)}}\}, \tag{8}$$

where $W^{(l)}$ is a weights matrix at layer $(l)$, and is distinctive among the ensuing expressions. For the transformation of $A$, only the LayerNorm is used:

$$H_A^{(0)} = \underbrace{\text{LayerNorm}\left(\mathbf{A}_{ij}\right),}_{h_{A_{ij}}} \qquad \forall\ (i, j) \in A. \tag{9}$$

The authors apply a custom graph convolution to the produced the latent vectors in two phases. In the first phase, they convolve over the triplet $(H_x, H_A^T, H_\delta)$ to compute Message $m_{(ij)}^{(l)}$, which is the convolution on the constraints side, that is represented as $C_{x\text{-}\delta}$. This convolution is calculated as follows:

$$h_{ij}^{(l)} = W^{(l)}h_{x_i}^{(l-1)} + W^{(l)}h_{A_{ij}}^{T\ (l-1)} + W^{(l)}h_{\delta_j}^{(l-1)},$$
$$\bar{h}_{ij}^{(l)} = \text{LayerNorm}\left(W^{(l)}\left(\sigma\left(W^{(l)}h_{ij}^{(l)}\right)\right)\right).$$

In this step, the effect of each variable on the constrained regionis learned. The AGGREGATE operation is defined as reduction by summation of all normalized features from the neighboring nodes in the set $\delta$, which is referred to as Message $m_{ij}$:

$$\text{AGGREGATE}^{(l)}(H_x, H_A, H_\delta) = \underbrace{\sum_{j\in\mathcal{N}(i)} \bar{h}_{ij}^{(l)}}_{m_{ij}^{(l)}}, \tag{10}$$
$$C_{x\text{-}\delta}^{(l)} = \text{UPDATE}^{(l)}(\{i \in |H_x|\}) = W^{(l)}h_{x_i} + W^{(l)}\left(m_{ij}^{(l)}\right).$$

$\mathcal{N}(i) = \{j | A_{ij} \neq 0\ \forall\ j \in |A_i|\}$ is the set of neighboring nodes and $|\cdot|$ is the cardinality operator. In the second phase, the same operations are used . However, to convolve over the triplet $(C_{x\text{-}\delta}, A, x)$, a similar concept of learning is performed from the side of the constraints, $C_\delta$, against the previously produced variable convolution. This process is defined as follows:

$$C_{\delta\text{-}x}^{(l)} = \text{UPDATE}^{(l)}(\{i \in |H_\delta|\}) = W^{(l)}h_{\delta_i} + W^{(l)}\left(C_{x\text{-}\delta}^{(l)}\right). \tag{11}$$

Eventually, log probabilities over possible branching candidates are estimated by a nonlinear transformation of the resulting convolution $C_{\delta\text{-}x}$, and the branching candidate has the highest log probability:

$$\hat{v} = \text{argmax}\left(W^{(l+1)}\left(\sigma\left(W^{(l+1)}C_{\delta\text{-}x}^{(l)}\right)\right)\right). \tag{12}$$

### *4.2. GraphSAGE*

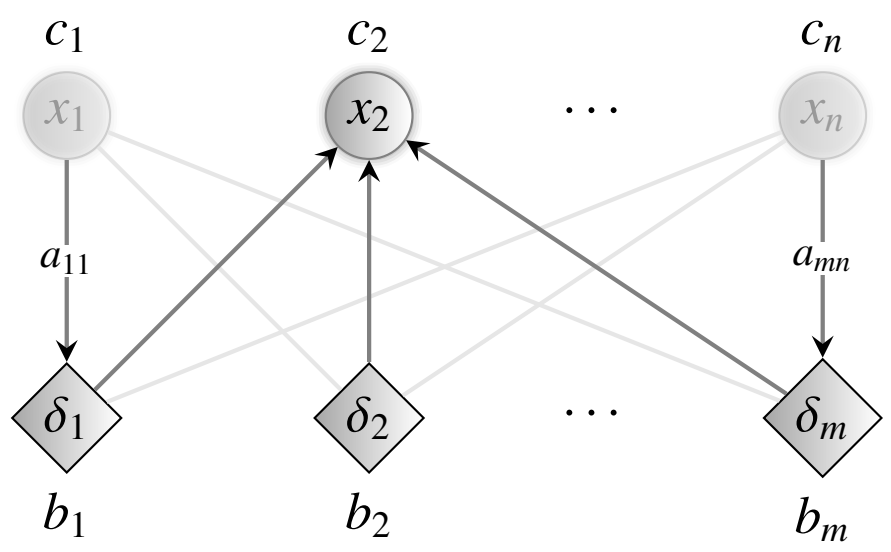

Figure 5: GraphSAGE accumulates information from a selection of neighboring nodes, with more distant nodes becoming increasingly influential in the aggregated information as the process continues.

The inputs to the nonbipartite deep graph neural networks are curated differently; the constraints $\delta_{n\times 5}$ are first padded with zeros to match the variable row dimension, resulting in $\delta_{m\times 19}^{pad} = [\delta_{m\times 5}\ \mathbf{0}_{m\times 14}]$. Then, the variables $x_{n\times 19}$ and the padded constraints $\delta_{m\times 19}^{pad}$ are augmented to form the input $H = \begin{bmatrix} x_{n\times 19} \\ \delta_{m\times 19}^{pad} \end{bmatrix} \in \mathbb{R}^{(n+m)\times 19}$, which is then fed en bloc to the classifier. The adjacency matrix would then be a zero-diagonal matrix with the coefficients being off-diagonal[1]:

$$\mathbf{A} = \begin{array}{c} n\left\{ \vphantom{\begin{matrix} 0 \\ \vdots \\ 0 \end{matrix}} \right. \\ m\left\{ \vphantom{\begin{matrix} 0 \\ \vdots \\ 0 \end{matrix}} \right. \end{array} \begin{bmatrix} \overbrace{\begin{matrix} 0 & \cdots & 0 \\ \vdots & \ddots & \vdots \\ 0 & \cdots & 0 \end{matrix}}^{n} & \overbrace{\begin{matrix} a_{11} & \cdots & a_{1m} \\ \vdots & \ddots & \vdots \\ a_{n1} & \cdots & a_{nm} \end{matrix}}^{m} \\ \begin{matrix} a_{11} & \cdots & a_{1n} \\ \vdots & \ddots & \vdots \\ a_{m1} & \cdots & a_{mn} \end{matrix} & \begin{matrix} 0 & \cdots & 0 \\ \vdots & \ddots & \vdots \\ 0 & \cdots & 0 \end{matrix} \end{bmatrix}$$
$$= \begin{bmatrix} \mathbf{0} & A \\ A^T & \mathbf{0} \end{bmatrix}.$$

We start by taking a linear transformation of the inputs to produce $H' \in \mathbb{R}^{(n+m)\times 19\times d}$:

$$H'^{(0)} = \left[\underbrace{W^{(0)}x_1}_{h_1'^{(0)}}, \cdots, \underbrace{W^{(0)}x_n}_{h_n'^{(0)}}, \underbrace{W^{(0)}\delta_1}_{h_{n+1}'^{(0)}}, \cdots, \underbrace{W^{(0)}\delta_m}_{h_{n+m}'^{(0)}}\right], \tag{13}$$

[1]The matrix can be constructed and stored efficiently using coordinate format.

where $W^{(l)} \in \mathbb{R}^{d\times 19}$. The message produced for Node $i \in H'$, denoted $m_i$ ,is defined as:

$$m_i^{(l)} = \text{AGGREGATE}^{(l)}(h_i) = \underbrace{\frac{1}{|\mathcal{N}(i)|}\sum_{j\in\mathcal{N}(i)} h_j^{(l)}}_{\text{Neighborhood Averaging}}, \quad (14)$$

$$\text{UPDATE}^{(l)}(\{\forall\, i \in |H'^{(l)}|\}) = \sigma\begin{pmatrix} W^{(l)}h_i + \\ W^{(l)}m_i^{(l-1)}\end{pmatrix},$$

where $\mathcal{N}(i) = \{j\,|A_{ij} \neq 0, \forall j \neq i \in |H'|\}$. Log probabilities over the branching candidates are obtained by a linear transformation over the updated features. The branching candidate is then selected using:

$$\hat{v} = \text{argmax}\left(W^{(l+1)}H'^{(l)}\right) \quad (15)$$

### *4.3. Graph Attention Network (GAT)*

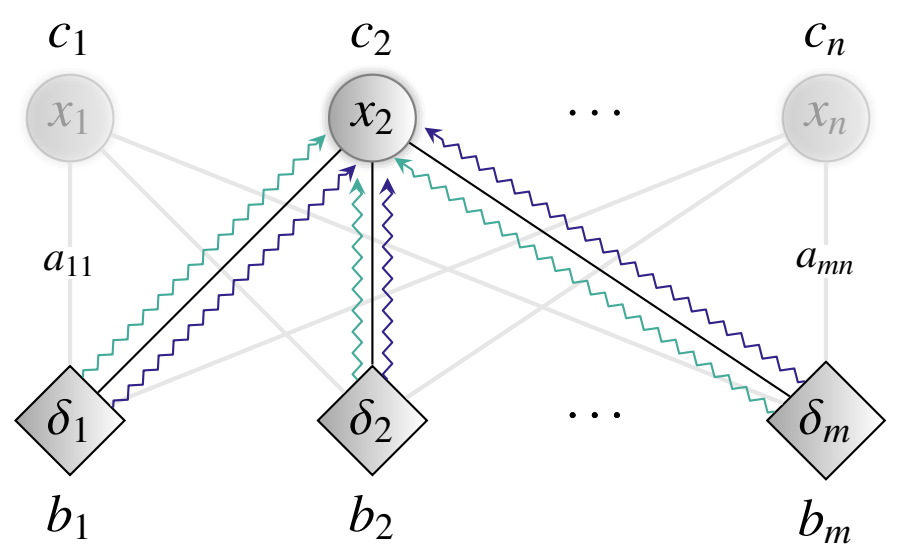

Figure 6: GAT assigns attention scores to the relationships between each node and its neighboring nodes. This is an example of a three-head attention mechanism.

The input to the GAT is the same as the input to GraphSAGE, $H$. Similarly, $H'$ is computed using Equation (13). We then train a self-attention mechanism, denoted as $a : \mathbb{R}^d \times \mathbb{R}^d \to \mathbb{R}$, defined as:

$$a^{(l)}(h'_i, h'_j) = \text{LeakyReLU}\left(w_a^{T(l)} \odot \left[h'_i \middle\| h'_j\right]\right),$$

where $\text{LeakyReLU}(y) = \max(0, y) + m \cdot \min(0, y)$, $w_a^{T(l)} \in \mathbb{R}^{2d}$, and $\|$ represents the concatenation operator. The attention score $\alpha_{ij}$ is then generated by normalizing the attention coefficients assigned to the neighboring nodes using $a$ with the softmax function and multiplying by $\mathbf{A}_{ij}$ in the following way:

$$\alpha_{ij}^{(l)} = \frac{\exp\left(a^{(l)}(h_i, h_j)\right)}{\sum_{j\in\mathcal{N}(i)}\exp\left(a^{(l)}(h_i, h_j)\right)} \odot \mathbf{A}_{ij}.$$

Such a score indicates the learned connectivity strength of Node $i$ to Node $j$ rescaled by the decision variable coefficient. The AGGREGATE operation performs a nonlinear combination of the learned attention scores and the linearly transformed inputs:

$$\text{AGGREGATE}^{(l)}(i) = \sigma\left(\sum_{j\in\mathcal{N}(i)} \alpha_{ij}h'_j\right).$$

The nonlinearity, $\sigma$, is used to prevent unintended fluctuations in the learned attention values. To integrate a $K$-head attention mechanism, for $K \in \mathbb{Z}^+$ (see Figure 6), the outputs are averaged among $K$ independent operators, and the message-protocol is constructed by:

$$m_i^{(l)} = \frac{1}{K_{(l)}}\left(\sum_{k=1}^{K_{(l)}} \text{AGGREGATE}^{(l)}(h_i)\right)$$
$$\text{UPDATE}^{(l)}(\{\forall\, i \in |H'^{(l)}|\}) = \sigma\begin{pmatrix} W^{(l)}h'_i + \\ W^{(l)}m^{(l-1)}(i)\end{pmatrix}. \quad (16)$$

The branching candidate is decided, analogously to GraphSAGE, by:

$$\hat{v} = \text{argmax}\left(W^{(l+1)}H'^{(l)}\right) \quad (17)$$

## 5. Experiments

The SB decision samples were drawn using École by formulating the CVRP and BPP as outlined in Sections 2.5 and 2.6, respectively. The CVRP training and evaluation instances were drawn from CVRPLIB (Lima et al. 2014), with six Instances from sets A and P (P et al. 1995) for training (A-n32-k5, A-n33-k5, A-33-k6, A-n39-k5, A-n44-k6, and P-n40-k5) and eight Instances from sets P, B, and M for evaluation: (P-n76-k5, P-n60-k15, P-n65-k10, B-n78-k10, B-n64-k9, B-n57-k7, M-n101-k10, and M-n151-k12). Each instance comprises of the following attributes: the number of customers, the demand of each customer, and the location of the depot and customers in a two-dimensional Euclidean space, where the cost $c_{ij}$ in (4a) corresponds to the Euclidean *distance* between node $i$ and node $j$.

Similarly, the BPP training instances were selected from (Beasley 2004), specifically sets U and T, with two instances for training (u100_00 and u80_00) and five instances for evaluation (t249_00, t120_00, t60_00, u250_00, and u500_00)

The training and evaluation instances are referred to as **tr** and **ts**, respectively. The configuring parameters in SCIP were set to their default, except for disabling the separation routine, enabling the search-tree-size profiling, and the branching strategy idempotence.

### *5.1. Training*

Due to the computational cost imposed by SB, we randomly set the strategy to either the RPC or SB at each branching. But, we only exported the branchings decided by the SB strategy to the training dataset. The size of which, denoted as $N$, is $10^5$ for each sampled dataset $\mathcal{D} = \{(x_i, A, \delta_i, v_i)\}_{i=1}^N$, where each record is an independent and identically distributed sample from the SB decisions.

The classifiers were trained to maximize the log likelihood (Pawitan 2013) using adaptive stochastic gradient ascent (Kingma and J. Ba 2014):

$$\mathcal{L}(\hat{v}, v) = \frac{1}{N}\sum_{i=1}^{N} \log \pi_\theta(\hat{v}_i = v_i | x_i, \delta_i). \quad (18)$$

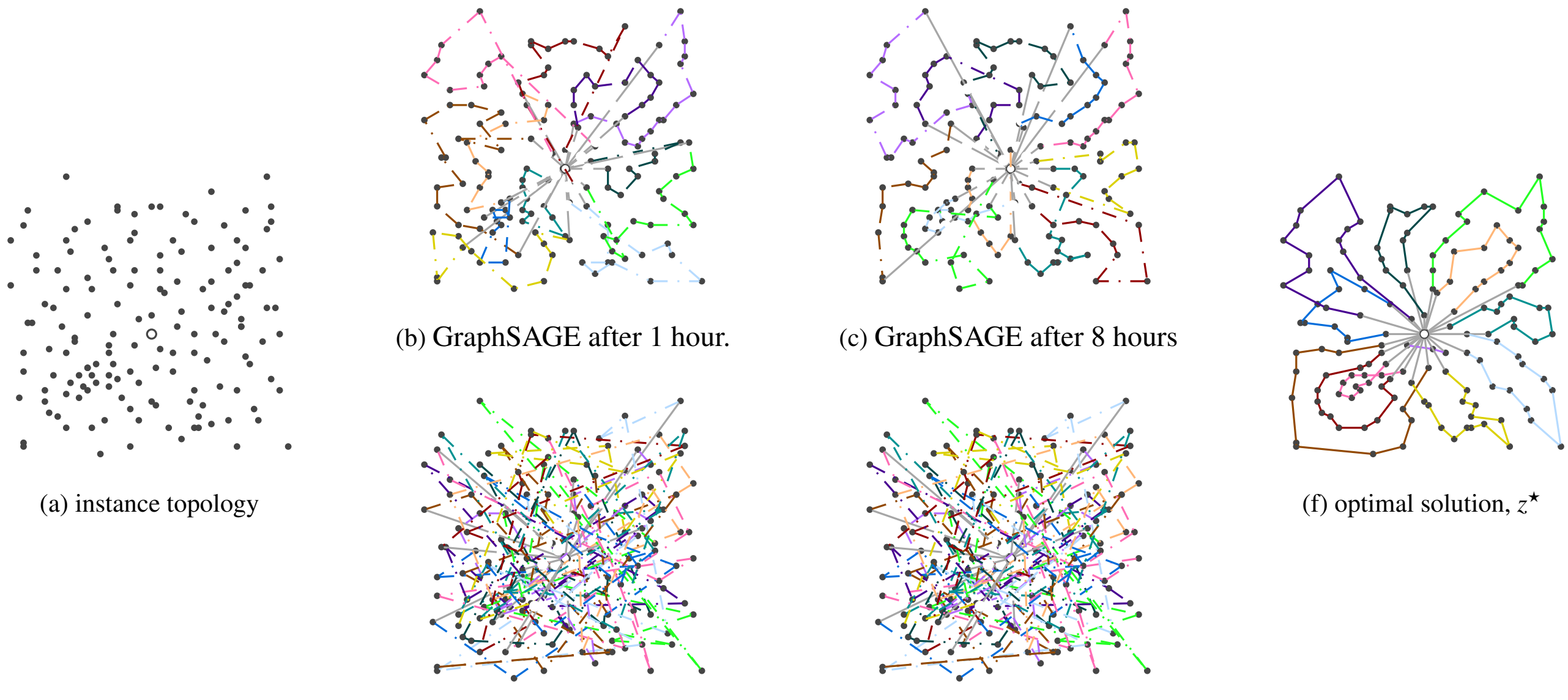

(a) instance topology

(b) GraphSAGE after 1 hour.

(c) GraphSAGE after 8 hours

(d) $\text{SCIP}_{\text{SB}}$ after 1 hour

(e) $\text{SCIP}_{\text{SB}}$ after 8 hours

(f) optimal solution, $z^\star$

Figure 7: Example on solving Instance M-n151-k12 using both GraphSAGE and SCIP with allowed runtime of 8 hours.

A verbal explanation of equation (18) is: given a set of constraints $\delta_i$, the variable features $x_i$, and the decided branching variable $v_i$, we aim to maximize the likelihood that a classifier, or policy, $\pi$ parameterized by the weights $\theta$ selects the same branching decision, $\hat{v}_i$, using either (12), (15) or (17).

At each iteration, we tested the performance on a sequestered validation set. We did not use dropout or add self-loops. The tuning parameters of each classifier are set to the following values:

- GCNN: $d = 64$, $l = 1$, $m = 0.2$;
- GraphSAGE: $d = 64$ and $l = 5$;
- GAT: $d = 128$, $l = 2$, $K_1 = 2$, and $K_2 = 4$.

These values were chosen through an experiential process that targeted minimizing the empirical *training* error (to prevent an overuse of the validation sets (Hastie, Tibshirani, and Friedman 2009; Feldman, Frostig, and Hardt 2019; Mania et al. 2019)).

### 5.2. Evaluation

We evaluated five branching strategies: SB, RPC (which are the inherently built branching strategies), GCNN, GraphSAGE, and GAT (which are the learned strategies) using SCIP with the time limit set to 1 hour, 2 hours, 4 hours, or 8 hours when solving each evaluation instance $j \in \textbf{ts}$. For brevity, the gap achieved by SCIP using the SB or RPC strategies will be referred to as $\text{SCIP}_{\text{SB}}$ and $\text{SCIP}_{\text{RPC}}$, respectively. We also use the notation $\text{SCIP}_s$ to refer to either of them. In Figure 7, we contrast the solution quality obtained by GraphSAGE against that of $\text{SCIP}_{\text{SB}}$ on the first (see Figures 7b and 7d) and eighth (see Figures 7c and 7e) hours when solving the CVRP instance M-n151-k12 (see Figure 7a). As shown, there is a strong resemblance between the constructed tours and the optimal ones (see Figure 7f) only after a single hour of execution. However, we shall introduce solid assessment metrics.

We define the optimality gain using a four-faceted comparison. Let $\Delta f_{j,t}^{\text{SCIP}_s}$ be the gap of SCIP using the strategy $s$ on $\textbf{ts}_j$, which is the $j$-th instance of $\textbf{ts}$, after $t$ hours of solving. We let $\Delta f_{j,t}^{m_i}$ be the gap of model $m$ when trained on $\textbf{tr}_i$, which is the $i$-th instance of $\textbf{tr}$, and evaluated on $\textbf{ts}_j$ after $t$ hours of solving. Then, we assess the quality of our proposed approach by calculating the following relative performance measure:

$$\Delta f_{\text{SCIP}_s}^{m}(i, j, t, k) = \frac{\Delta f_{j,t}^{m_i}}{\Delta f_{j,(2^k t)}^{\text{SCIP}_s}}, \tag{19}$$

where we chose $k = 0, \ldots, 3$, $t \in T = \{1, \ldots, 2^{3-k}\}$. The measure (19) expresses the ratio between the gap of two solutions: the first solution is the proposed approach, using model $m$, trained on $\textbf{tr}_i$, evaluated on instance $\textbf{ts}_j$, and the gap is reported after $t$ hours; the second solution is the one provided by the solver SCIP, using SB or RPC, and evaluated on the same instance $\textbf{ts}_j$, and the gap is reported after $2^k t$ hours. The elongation factor $2^k$ bestows upon SCIP a better chance to find a lower gap. When not found, it would assert the capability of our approach. For example, when $k = 0$, the duality gap of each solution is reported at $\{1, 2, 4, 8\}$ to examine a contrast of the classifiers' performance against $\text{SCIP}_s$ under *equal* time limits, denoted (1:1). When $k = 1$, the gap is reported at $\{1, 2, 4\}$ and $\{2, 4, 8\}$, which yields a comparison of performance with *twice* the time limit for $\text{SCIP}_s$, denoted (1:2). Two further comparisons are made at (1:4) and (1:8), where $k = 2$ and $k = 3$, respectively, to determine if even larger gains are achieved by increasing the time limit by *four* and *eight* times.

For example, when $k = 0$, the duality gap of each solution is reported at $\{1, 2, 4, 8\}$ to examine a contrast of the classifiers' performance against $\text{SCIP}_s$ under *equal* time limits, denoted (1:1). When $k = 1$, we report the gap at $\{1, 2, 4\}$ and $\{2, 4, 8\}$, which yields a comparison of performance with twice the time limit for $\text{SCIP}_s$, denoted (1:2). We include two further comparisons at (1:4) and (1:8), where $k = 2$ and $k = 3$, respectively, to determine if our approach can achieve higher gains by increasing the time limit by four and eight times for $\text{SCIP}_s$.

We evaluate the average performance of each classifier in two ways: (1) by fixing the training instances and calculating the mean performance of *each* time-window comparison as previously described across the evaluation instances, using the following equation:

$$\Delta f^m_{\text{SCIP}_s}(i, t, k) = \frac{1}{|\mathbf{ts}|} \sum_{j \in \mathbf{ts}} \Delta f^m_{\text{SCIP}_s}(i, j, t, k). \tag{20}$$

(2) by averaging the performance across both the training and evaluation instances using the following equation (i.e., reduction by $i$ and $j$):

$$\Delta f^m_{\text{SCIP}_s}(t, k) = \frac{1}{|\mathbf{tr}|} \sum_{i \in \mathbf{tr}} \Delta f^m_{\text{SCIP}_s}(i, t, k). \tag{21}$$

We also report two more measures to assess the aptitude of the classifiers to generalize in performance to larger and more complex instances, which are the GED and TSE. The GED is the number of edits required to make the topology of a graph $\mathbf{tr}_i$ isomorphic to the topology of a graph $\mathbf{ts}_j$. The GED provides insights into the topological similarities between instances $i$ and $j$. SCIP calculates the TSE as the relative increase in nodes between two consecutive tree states among all explored paths. The TSE of a classifier $m$ trained on instance $i$ and used as a branching strategy to solve evaluation instance $j$ for $t$ hours is denoted as $\text{TSE}^{m_i}_{j,t}$, while the TSE of the standard branching strategy $s$ is denoted as $\text{TSE}^{\text{SCIP}_s}_{j,t}$. The ratio between these two quantities calculates how many more nodes the learned strategy needs to explore on larger instances compared to the standard one, $s$:

$$\Delta\text{TSE}^{m_i}_{\text{SCIP}_s} = \frac{1}{|T|} \sum_t \left( \frac{\text{TSE}^{m_i}_{j,t}}{\text{TSE}^{\text{SCIP}_s}_{j,t}} \right) \tag{22}$$

In Section 6, we normalized the displayed values of both the GED and $\Delta\text{TSE}^m_{\text{SCIP}_s}$ to the interval $[0, 1]$ for convenience.

## 6. Results

We present the optimality gain described by (19) in Figures 8 and 9 for the CVRP and Figures 12 and 13 for the BPP. Since the four figures encompass the same elements. We will only describe the elements of Figures 8 and 9 and convey the same theme to the ones for the BPP. Each of Figures 8 and 9, which details the comparison against $\text{SCIP}_{\text{SB}}$ and $\text{SCIP}_{\text{RPC}}$, comprises six rows for the training instances and three columns for the classifiers. We will use the notation $\text{Fig}(s, m, i)$ to refer to the figure describing the results of classifier $m$ when trained on instance $i$ and evaluated against $\text{SCIP}_s$ on all instances $\mathbf{ts}$; e.g., Fig(RPC, GCNN, A-n32-k5) is the top and leftmost sub-figure in Figure 9, representing the performance gain of GCNN when trained on Instance A-n32-k5 and compared against $\text{SCIP}_{\text{RPC}}$. Each $\text{Fig}(s, m, i)$ has eight interior columns, one for each $\mathbf{ts}_j$, denoted $\text{Fig}(s, m, i, j)$; each is a lump of *four* distinctive gray-level bars to *visually* differentiate the four elongation factors $2^k$ defined in (19), and each is denoted $\text{Fig}(s, m, i, j, k)$.

For example, the first bar is $\text{Fig}(s, m, i, j, 0)$ and has four symbols − + □ ○ to represent $\Delta f^m_{\text{SCIP}_s}(i, j, t, 0)$ at $t \in T = \{1, \dots, 2^3\}$. That is, a performance ratio of the learned branching strategies to the standard ones when both are given equal time limits, which are 1, 2, 4, and 8 hours. The second bar is $\text{Fig}(s, m, i, j, 1)$ and has three symbols + □ ○ to represent $\Delta f^m_{\text{SCIP}_s}(i, j, t, 1)$ at $t \in T = \{1, \dots, 2^2\}$. The other two bars follow a similar pattern for the remaining fourfold (1:4) and eightfold (1:8) time limit extensions for $\text{SCIP}_s$. In a limited number of experiments, $\text{SCIP}_s$ fails to obtain a feasible solution in a certain time limit $t$ on an evaluation instance $j$; we distinguish those observations by using the same symbols as described previously, however, capped by an Ω, e.g., and , and shifted to the top of the bar $\text{Fig}(s, m, i, j, k)$. A single Ω spanning the entire column $\text{Fig}(s, m, i, j)$ indicates the same meaning across the different values of $k$, and accordingly, $t$. Equivalently, we use a flipped ℧ to designate the similar behavior for a classifier $m$. Additionally, for each $\text{Fig}(s, m, i)$, there are two stacked lines on the top: the dotted line shows the topological *complexity* of each evaluation instance relatively to the training instance, computed by the GED; the triangle-dashed line reflects the $\Delta\text{TSE}^{m_i}_{\text{SCIP}_s}$ (22). The pale-red area height in $\text{Fig}(s, m, i, j)$, , highlights the standard deviation of the average gains achieved by classifier $m$ trained on instance $j$ against $\text{SCIP}_s$ over the different values $k$

If a $\text{Fig}(s, m, i, j)$ is rendered in light blue color, , it indicates the highest performance gain when solving instance $j$ using classifier $m$ trained on instance $i$ against $\text{SCIP}_s$, e.g., Fig(RPC, GCNN, A-n33-k5, P-n65-k10) The dashed line, , fixed at $\Delta f^m_{\text{SCIP}}(i, j, t, k) = 1$ in a $\text{Fig}(s, m, i)$ separates the results into two segments. The area below the line is of particular interest because it is the region where the performance of a classifier $m$ trained on instance $i$ is equal to or better than that of $\text{SCIP}_s$ for all instance $j \in \mathbf{ts}$.

### 6.1. *Findings*

#### 6.1.1. *CVRP*

When examining Figure 8, which displays the classifiers' performance compared to $\text{SCIP}_{\text{SB}}$, it is visible that the trained classifiers outperform $\text{SCIP}_{\text{SB}}$ in most experiments. They either meet or improve upon $\text{SCIP}_{\text{SB}}$'s performance, as indicated by the dashed line representing equal performance. These improvements are most palpable in Fig(SB, GCNN, A-n33-k6, P-n76-k5) and Figure (SB, GraphSAGE, A-n33-k6, M-n151-k12). Additionally, $\text{SCIP}_{\text{SB}}$ was unable to find a feasible solution within the first hour for Instance P-n76-k5, as shown in Fig(SB, m, i, P-n76-k5, 0), or within the first, second, and fourth hour for Instance B-n57-k7, as shown in Fig(SB, m, i, B-n57-k7, 0). This trend continues for Instance B-n64-k9 across all time windows, as indicated by the Ω symbol. No-tably, the three classifiers find significantly lower gaps on Instance M-n151-k12, even when given the maximum time allowance to $\text{SCIP}_{\text{SB}}$ (i.e., eight hours). Unfortunately, the GED and $\Delta\text{TSE}^m\text{SCIP}_{\text{SB}}$ showed fluctuations with no apparent correlation to the classifiers' performance, making it challenging to affirm generalization capabilities. When comparing the performance of the individual classifiers, GCNN achieves the lowest gap in four out of eight instances; when compared to GraphSAGE, the surplus is, however, modest. GAT is able to outperform the other two classifiers at the eighth hour, which the average-performance figures emphasize. It is also worth mentioning that the classifiers trained on Instance A-n33-k6 found the lowest gap on Instances P-n76-k5, B-n78-k10, and M-n151-k12, which we attribute to a possible similarity of the complexity among them.

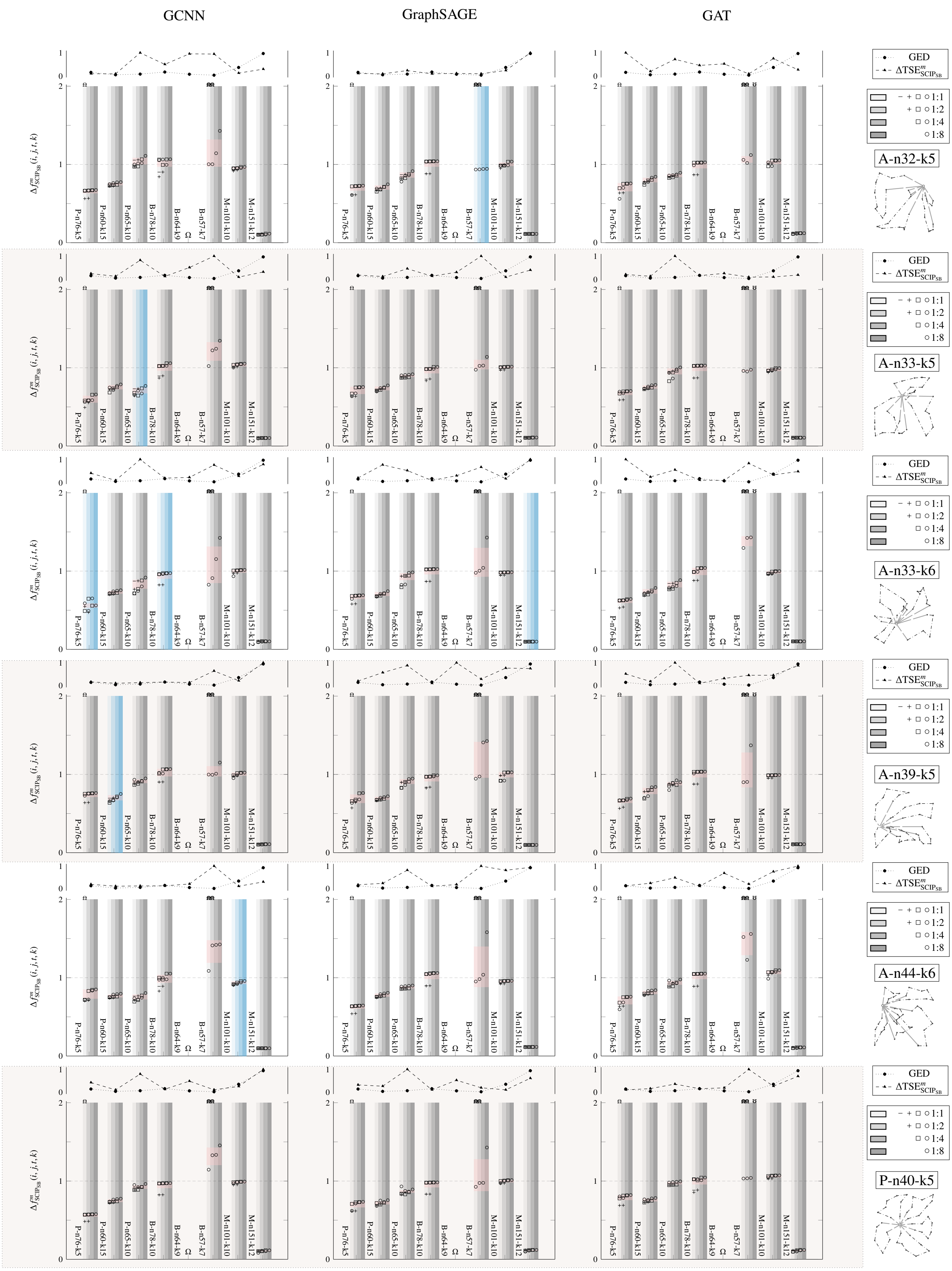

Figure 8: The performance of the three CVRP classifiers against $SCIP_{SB}$ calculated by (19), across the training instances.

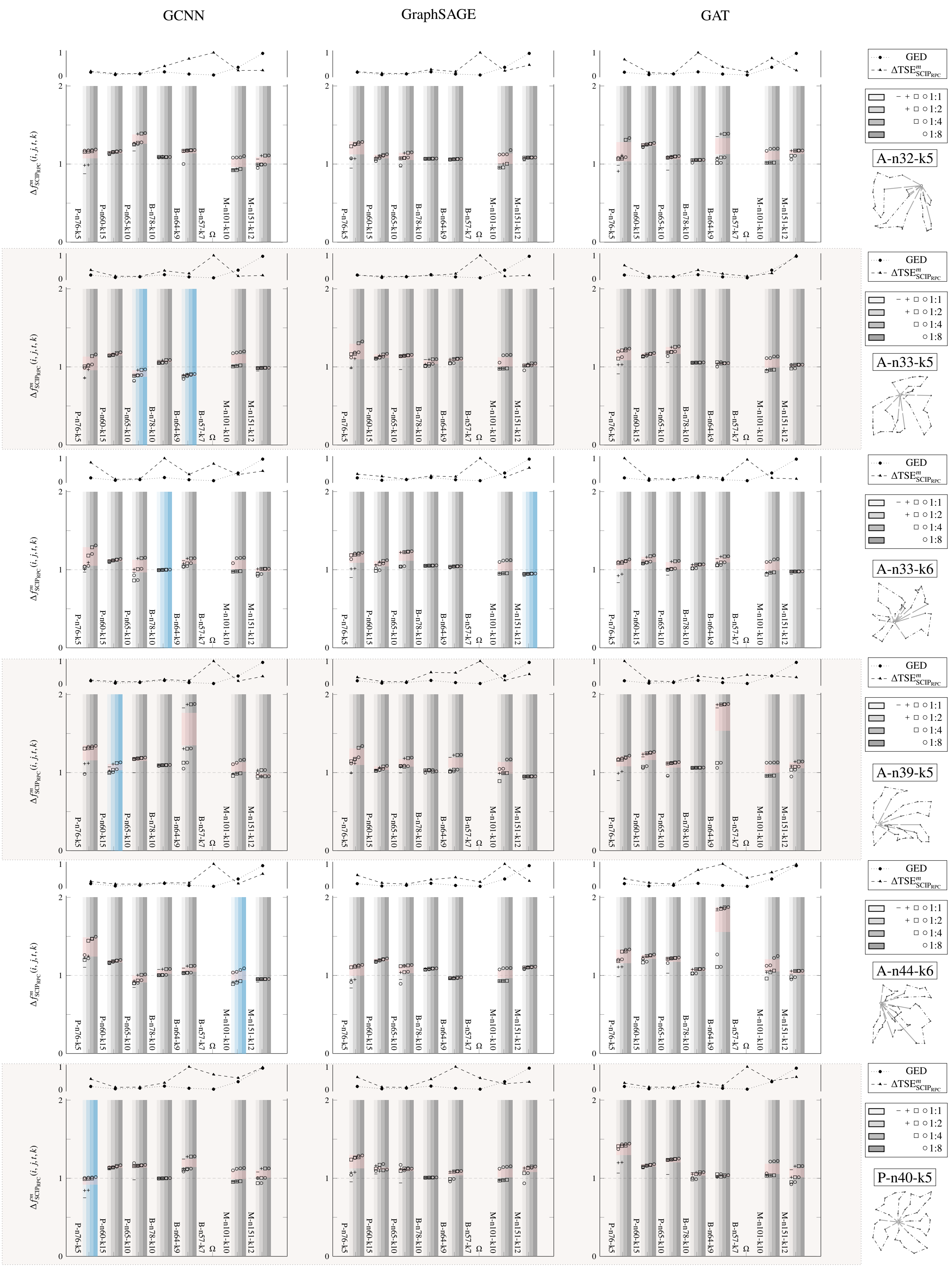

Figure 9: The performance of the three CVRP classifiers against $SCIP_{RPC}$ calculated by (19), across the training instances.

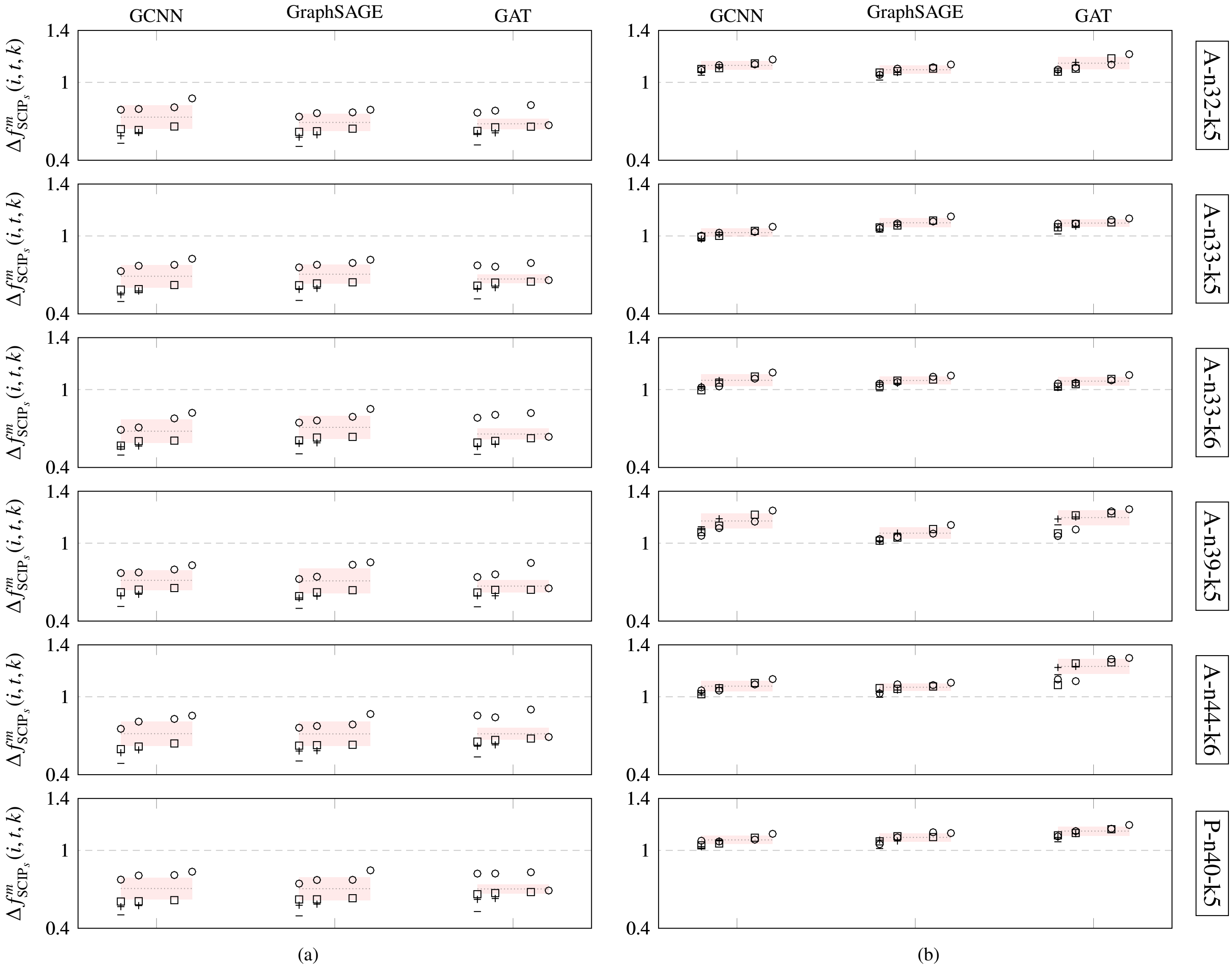

Figure 10: The average performance of each classifier across the CVRP evaluation instances; (a) compares against $\text{SCIP}_{\text{SB}}$ and (b) compares against $\text{SCIP}_{\text{RPC}}$.

The classifiers' performance gains are also reflected in Figure 9, comparing against $\text{SCIP}_{\text{RPC}}$. Using the RPC strategy improves SCIP's performance. Nonetheless, at least one of the classifiers achieves equivalent performance with slight improvement or decline in at least four hours less. The GCNN remains the top-performing classifier on five out of eight evaluation instances. The number of the explored nodes is significantly smaller than those of $\text{SCIP}_{\text{RPC}}$, this is illustrated the triangle-dashed line representing the value of $\Delta\text{TSE}^m_{\text{SCIP}_{\text{RPC}}}$. The fourth bar shows particularly intriguing results, depicting the performance of the classifiers against a time allowance of an eightfold increase for $\text{SCIP}_s$. The potential gains when using the standard branching strategy are minute, and in some cases, there might be a net loss, which might not be worthwhile in time-critical sectors. On average, the classifiers consistently find lower gaps than $\text{SCIP}_{\text{SB}}$ in considerably less time, as depicted in Figure 10a, which is calculated using (20). Although $\text{SCIP}_{\text{RPC}}$ improves upon $\text{SCIP}_{\text{SB}}$'s performance as depicted in Figure 10b, the classifiers still exhibit impressive time savings. We further support these conclusions by Figure 11, which is calculated using (21), demonstrating the classifiers' capability to generalize to supposedly more complex instances[2]. However, we refrain from using this observation as a response to the research question (C), given the unconfirmed correlation of the $\Delta\text{TSE}^{m}_{\text{SCIP}_s}$ or the GED with the classifiers' obtained gap.

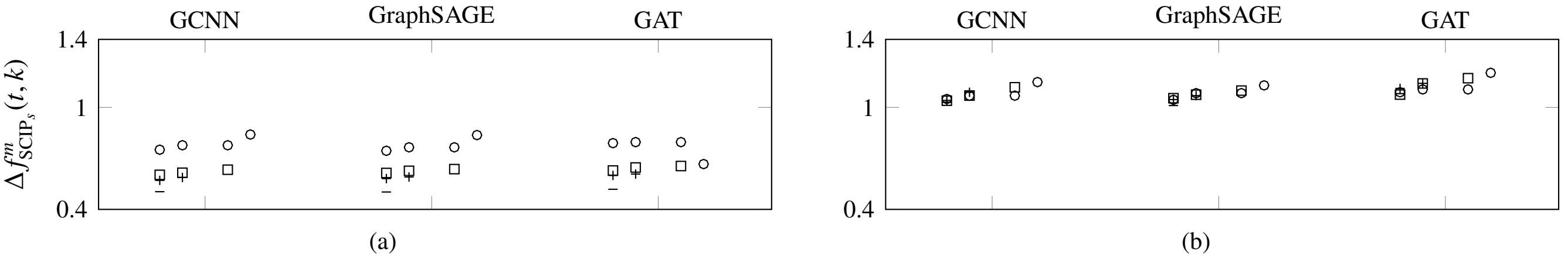

Figure 11: The performance of each classifier averaged by both of the CVRP training and evaluation instances against (a) $\text{SCIP}_{\text{SB}}$ and (b) $\text{SCIP}_{\text{RPC}}$.

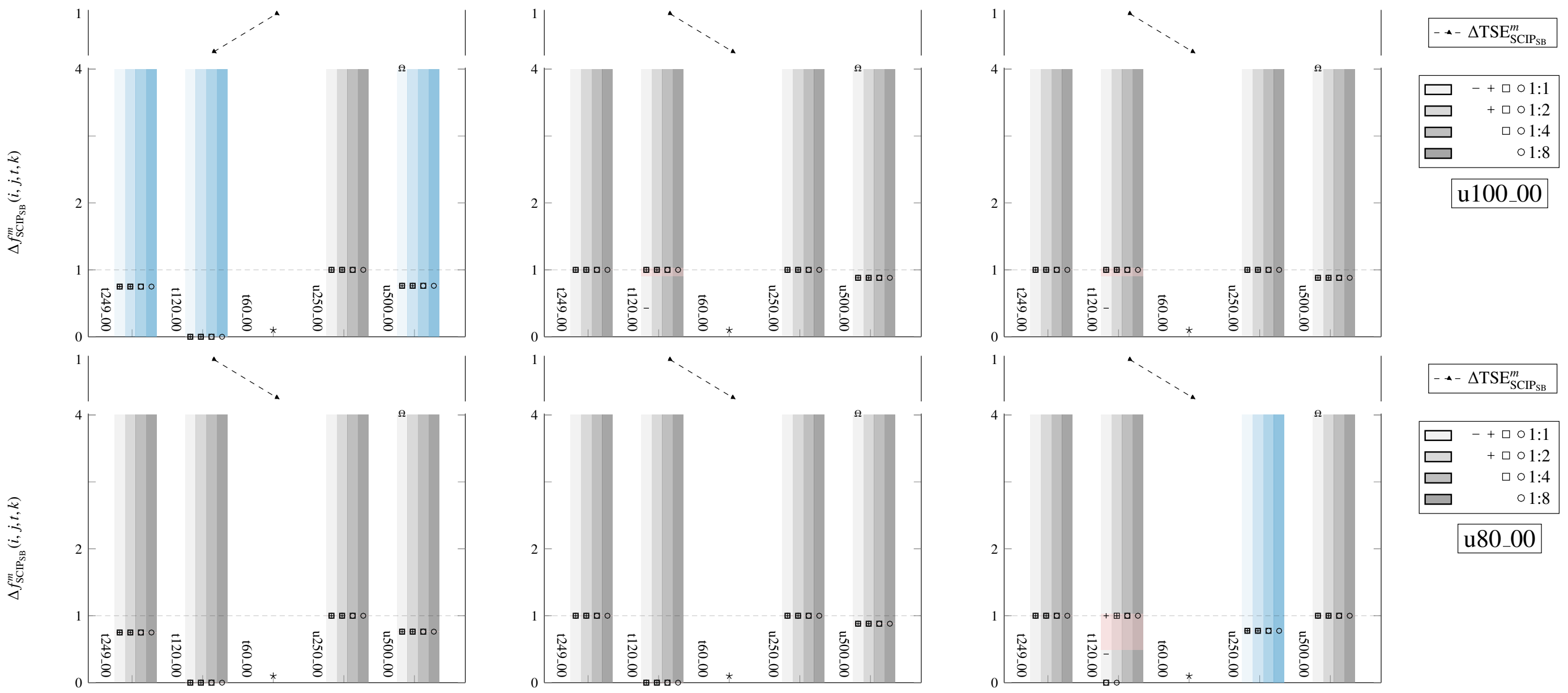

Figure 12: The performance of the three BPP classifiers against $\text{SCIP}_{\text{SB}}$ calculated by (19), across the training instances.

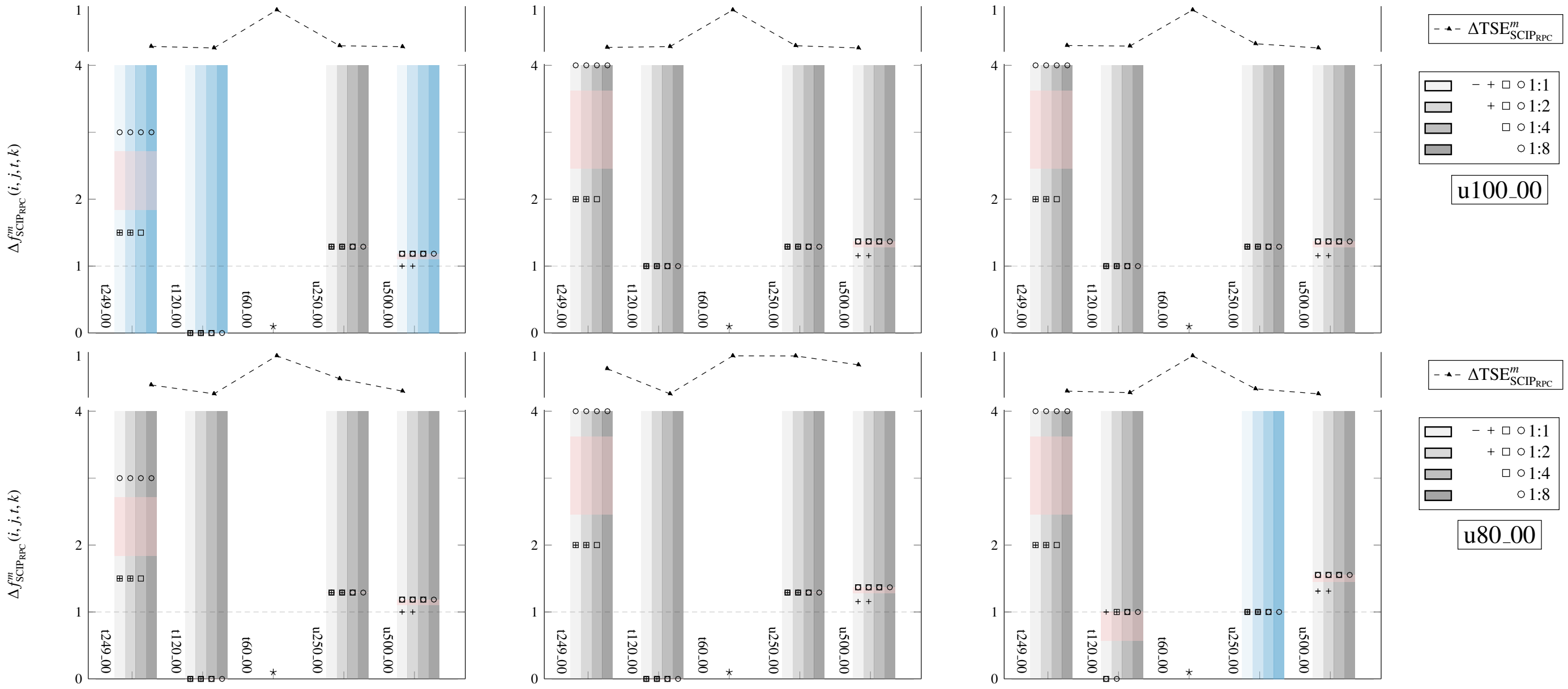

Figure 13: The performance of the three BPP classifiers against $\text{SCIP}_{\text{RPC}}$ calculated by (19), across the training instances.

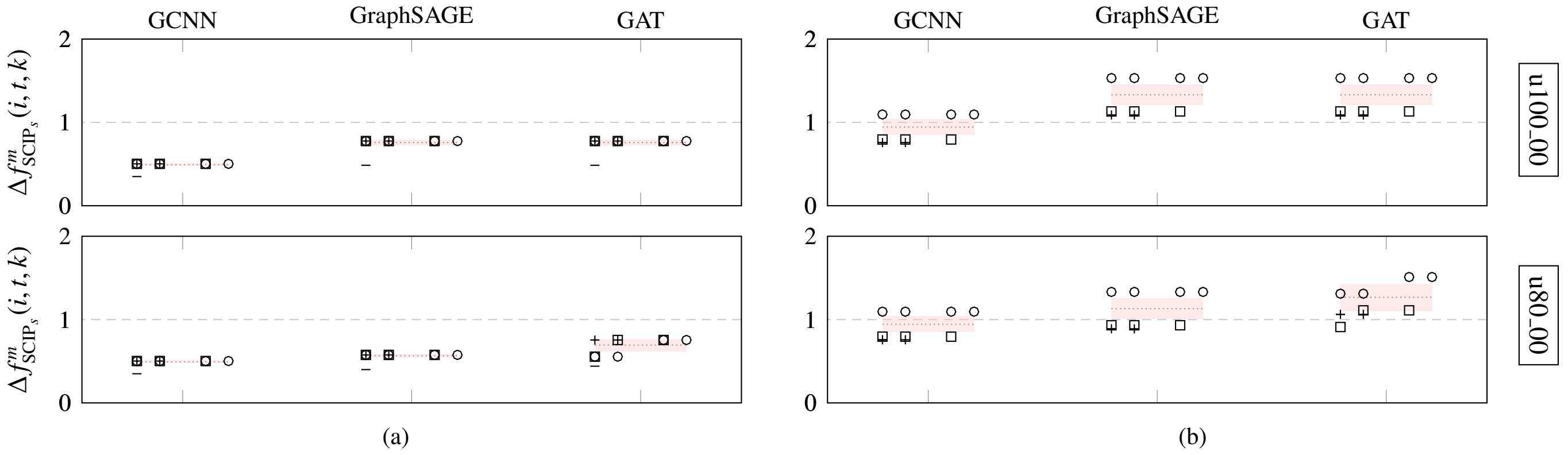

Figure 14: The average performance of each classifier across the BPP evaluation instances; (a) compares against $\text{SCIP}_{\text{SB}}$ and (b) compares against $\text{SCIP}_{\text{RPC}}$.

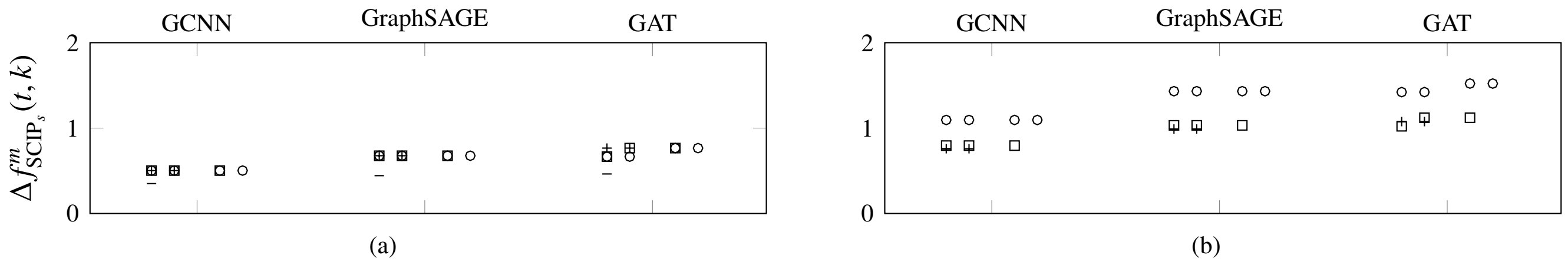

Figure 15: The performance of each classifier averaged by both of the BPP training and evaluation instances against (a) $\text{SCIP}_{\text{SB}}$ and (b) $\text{SCIP}_{\text{RPC}}$.

### *6.1.2. BPP*

Similar improvements against $\text{SCIP}_{\text{SB}}$ are apparent in Figure 12. However, we noticed a pattern change; for example, the extended time limit no longer allows for further duality gap optimization, either for $\text{SCIP}_{\text{SB}}$ or for classifier $m_i$, which we observed by the consistent values of $\Delta f^{m}_{\text{SCIP}_s}(i, j, t, k)$ across varying values of $k$. This is also asserted by the absence of tree-size estimates for three out of five evaluation datasets: t249_00, u250_00, and u500_00, due to the stagnant state of the search tree. We suspect that this is due to obtaining suboptimal distribution of items to bins within a limited time frame and that more time would be required to find improved solutions. Nonetheless, both $\text{SCIP}_{\text{SB}}$ and the classifiers solved dataset t60_00 optimally; however, $\text{SCIP}_{\text{SB}}$ failed to find a feasible solution for dataset t120_00. In Figure 13, $\text{SCIP}_{\text{RPC}}$ appears to have an advantage on larger time windows for dataset t249_00, with a significantly lower gap than that of any of the classifiers. It also successfully obtains a feasible solution for dataset t120_00, albeit with a performance lead over the classifiers, which have solved the problem optimally. All of $\text{SCIP}_{\text{SB}}$ and $\text{SCIP}_{\text{RPC}}$, as well as GCNN, GraphSAGE, and GAT, found optimal solutions for dataset t60_00 within the first hour. On average, the classifiers consistently achieve lower gaps than $\text{SCIP}_{\text{SB}}$ and show comparable performance to $\text{SCIP}_{\text{RPC}}$, as shown in Figures 14 through 15. It is also worth noting that variations in the training instances have minimal impact on the quality of the obtained solutions.

## 7. Conclusion

Our study underscores the potential of three geometric deep neural networks—graph convolutional neural network, GraphSAGE, and graph attention network—to either surpass or closely match the performance of the Strong Branching strategy in tackling the intricacies of the capacitated vehicle routing problem (CVRP) and bin packing problem. These classifiers exhibit minimal generalization error when confronted with new and more complex instances. Nonetheless, it is apparent that the existing literature lacks comprehensive metrics for accurately assessing the complexity of integer programs, a challenge attributed to their inherent intricacies.

While our findings are encouraging, they also point to challenges when dealing with larger instances with a significant number of customers. These challenges encompass issues related to generating decision samples for effective SB training of custom classifiers, as well as the current approach's limitations in devising strategies that surpass those it was trained on. To address these gaps, we propose two potential directions:

Firstly, leveraging the capabilities of UG (Shinano et al. 2012), a parallelized implementation of SCIP, that offers a promising avenue for refining our sampling tools. Additionally, initializing the training process with learned weights from smaller CVRP instances or different problem domains could potentially enhance classifier accuracy and efficiency when applied to larger instances.

[2]We excluded the instances that $\text{SCIP}_s$ was unable to find a feasible solution for.

Secondly, our ongoing work centers on reimagining the CVRP as a hide-and-seek game (J. von Neumann and Morgenstern 1947; John von Neumann 1953) involving two reinforcement-learning agents. In this paradigm, the objective is to maximize the *negative* relative primal-dual gap (Bichler et al. 2021). This approach involves one agent representing "hiding" elements (analogous to customers) and another agent as the "seeker" (equivalent to vehicles) in the pursuit of optimal solutions. The quality of the training set is no longer a limitation on this method.

In conclusion, this study uncovers promising avenues for the application of deep neural networks in addressing complex combinatorial optimization challenges, while also highlighting the need for further research in refining and expanding these approaches for substantial cases.

## 8. Acknowledgment

We express our gratitude for the partial support extended through the National Research Council of Canada's Artificial Intelligence for Logistics Program under Grant A-0035517. Furthermore, we acknowledge the contribution of the Digital Research Alliance of Canada for providing us with the computing hardware essential for the execution of our experiments. Thanks are also due to Brendan O'Donoghue for his valuable suggestions during the research process.